\documentclass{article}
\PassOptionsToPackage{numbers,sort, compress}{natbib}
\usepackage[final]{neurips_2020}
\usepackage[utf8]{inputenc}
\usepackage[T1]{fontenc}
\usepackage{amsfonts}
\usepackage{amsmath}
\usepackage{accents}
\usepackage{amssymb}
\usepackage{bbm}
\usepackage{bm}
\usepackage{booktabs}
\usepackage{chngcntr}
\usepackage{epsfig}
\usepackage{fontawesome}
\usepackage{graphicx}
\usepackage{mathtools}
\usepackage{nicefrac}
\usepackage{pifont}
\usepackage{subcaption}
\usepackage{tabularx}
\usepackage{todonotes}
\usepackage{url}
\usepackage{wrapfig}
\usepackage{xcolor}
\usepackage{hyperref}
\usepackage[capitalise,nameinlink]{cleveref}
\makeatletter
\newcommand{\cmark}{\ding{51}}%
\newcommand{\xmark}{\ding{55}}%
\makeatother
\newcommand{\bx}{\bm{x}}
\newcommand{\bz}{\bm{z}}

\newcommand{\ie}{\textit{i}.\textit{e}.}
\newcommand{\eg}{\textit{e}.\textit{g}.}
\newcommand{\ul}[1]{\underline{{#1}}}
\newcommand\methodname{SeLaVi}
\def\optrow#1\\{} %
\def\labelswitch#1{\label{#1}}
\def\dontshowthisinappendix#1{#1}
\def\showthisinappendix#1{}

\makeatletter
\renewcommand{\paragraph}{%
\@startsection{paragraph}{4}%
{\z@}{0.25em}{-1em}%
{\normalfont\normalsize\bfseries}%
}
\makeatother

\title{Labelling unlabelled videos \\ from scratch with multi-modal self-supervision}
\author{
  Yuki M. Asano$^{1}$\thanks{Joint first authors} \qquad
  Mandela Patrick$^{1,2}$\footnotemark[1] \qquad Christian Rupprecht$^1$ \qquad \textbf{Andrea Vedaldi}$^{1,2}$ \vspace{1mm}
\\
   $^1$ Visual Geometry Group, University of Oxford \\ \texttt{yuki@robots.ox.ac.uk} \\
   $^2$ Facebook AI Research \\  \texttt{mandelapatrick@fb.com}\vspace{1mm}\\
}

\begin{document}
\maketitle
\begin{abstract}
A large part of the current success of deep learning lies in the effectiveness of data -- more precisely: labelled data. Yet, labelling a dataset with human annotation continues to carry high costs, especially for videos. While in the image domain, recent methods have allowed to generate meaningful (pseudo-) labels for unlabelled datasets without supervision, this development is missing for the video domain where learning feature representations is the current focus. In this work, we a) show that unsupervised labelling of a video dataset does not come for free from strong feature encoders and b) propose a novel clustering method that allows pseudo-labelling of a video dataset without any human annotations, by leveraging the natural correspondence between the audio and visual modalities. An extensive analysis shows that the resulting clusters have high semantic overlap to ground truth human labels. We further introduce the first benchmarking results on unsupervised labelling of common video datasets Kinetics, Kinetics-Sound, VGG-Sound and AVE\footnote{Code will be made available at \url{https://github.com/facebookresearch/selavi}}. 
\end{abstract}

\raggedbottom

\section{Introduction}

One of the key tasks in machine learning is to convert continuous perceptual data such as images and videos into a symbolic representation, assigning discrete labels to it.
This task is generally formulated as \emph{clustering}~\citep{hartigan1972direct}.
For images, recent contributions such as~\citep{ji2018invariant,gansbeke2020learning,caron2018deep,asano2020self} have obtained good results by combining clustering and representation learning.
However, progress has been more limited for videos, which pose unique challenges and opportunities.
Compared to images, videos are much more expensive to annotate; at the same time, they contain more information, including a temporal dimension and two modalities, aural and visual, which can be exploited for better clustering.
In this paper, we are thus interested in developing methods to \textit{cluster video datasets without manual supervision}, potentially reducing the cost and amount of manual labelling required for video data.

Just as for most tasks in machine learning, clustering can be greatly facilitated by extracting a suitable \emph{representation} of the data.
However, representations are usually learned by means of manually supplied labels, which we wish to avoid.
Inspired by \citep{yan2020cluster}, we note that a solution is to consider one of the recent state-of-the-art self-supervised representation learning methods and apply an off-the-shelf clustering algorithm \emph{post-hoc}.
With this, we show that we can obtain very strong baselines for clustering videos.

Still, this begs the question of whether even better performance could be obtained by \emph{simultaneously learning to cluster and represent} video data.
Our main contribution is to answer this question affirmatively and thus to show that \textit{good clusters do not come for free from good representations}.

In order to do so, we consider the recent method SeLa~\citep{asano2020self}, which learns clusters and representations for still images by solving an optimal transport problem, and substantially improve it to work with multi-modal data.
We do this in three ways.
First, we relax the assumption made in~\citep{asano2020self} that clusters are equally probable; this is not the case for semantic video labels, which tend to have a highly-skewed distribution~\citep{gu2018ava,kinetics,youtube8m}, and extend the algorithm accordingly.
Second, we account for the multi-modal nature of video data, by formulating the extraction of audio and visual information from a video as a form of data augmentation, thus learning a clustering function which is invariant to such augmentations.
For this to work well, we also propose a new initialization scheme that synchronizes the different modalities before clustering begins.
This encourages clusters to be more abstract and thus `semantic' and learns a redundant clustering function which can be computed robustly from either modality (this is useful when a modality is unreliable, because of noise or compression).
Third, since clustering is inherently ambiguous, we propose to learn multiple clustering functions in parallel, while keeping them orthogonal, in order to cover a wider space of valid solutions.

With these technical improvements, our method for Self-Labelling Videos (SeLaVi) substantially outperforms the post-hoc approach~\citep{yan2020cluster}, SeLa~\citep{asano2020self} applied to video frames, as well as a recent multi-modal clustering-based representation learning method, XDC~\citep{alwassel2019self}.
We evaluate our method by testing how well the automatically learned clusters match manually annotated labels in four different video datasets: VGG-Sound~\citep{VGGSound}, AVE~\citep{tian2018ave}, Kinetics~\citep{kinetics} and Kinetics-Sound~\citep{arandjelovic17look}.
We show that our proposed model results in substantially better clustering performance than alternatives.
For example, our method can perfectly group 32\% of the videos in the VGG-Sound dataset and 55\% in the AVE dataset without using any labels during training.
Furthermore, we show that, while some clusters do not align with the ground truth classes, they are generally semantically meaningful (e.g.~they contain similar background music) and provide an interactive cluster visualization\footnote{\url{https://www.robots.ox.ac.uk/~vgg/research/selavi}}.

In a nutshell, our key contributions are:
\textbf{(i)} establishing video clustering benchmark results on four datasets for which labels need to be obtained in an unsupervised manner; \textbf{(ii)} developing and assessing several strong clustering baselines using state-of-the-art methods for video representation learning, and
\textbf{(iii)} developing a new algorithm tailored to clustering multi-modal data resulting in state-of-the-art highly semantic labels.

\section{Related work}

\paragraph{Unsupervised labelling for images.}

Early approaches to clustering images include agglomerative clustering~\citep{bautista2016cliquecnn} and partially ordered sets of hand-crafted features~\citep{bautista2017deep}, while more recent methods combine feature learning with clustering.
First, there are methods which propose to implicitly learn a clustering function by maximizing mutual information between the image and nuisance transformations~\citep{ji2018invariant,hu2017learning}.
Second, there are methods which use explicit clustering combined with representation learning~\citep{yang2016joint,caron2018deep,caron2019unsupervised,li2020prototypical,asano2020self,xie2016unsupervised, chang2017deep}. %
Lastly, there are methods which build on strong feature representations and, at a second stage, utilize these to obtain clusters~\citep{gansbeke2020learning,yan2020cluster,lee2019deep}.

\paragraph{Representation learning from videos.}

There is a growing literature on representation learning from videos.
Many of these methods are \emph{uni-modal}, leveraging works from the image domain~\citep{Chen2020ASF,asano2020a,tian2019contrastive,bachman2019learning,Wu_2018_CVPR,noroozi2017representation,gidaris2018unsupervised,gidaris2020learning,zhang2016colorful,noroozi2016unsupervised}, such as predicting rotations~\citep{jing2018self} and 3D jigsaw puzzles \citep{kim2019self}.
Other works leverage temporal information explicitly and predict future features~\citep{han2019video}, the order of frames~\citep{misra2016shuffle,lee2017unsupervised} and clips~\citep{clip_order}, the direction of time~\citep{wei2018learning} or the framerate~\citep{benaim2020speednet, cho2020selfsupervised}.
However, videos usually contain multiple modalities, such as audio, speech and optical flow.
\emph{Multi-modal} learning, originally proposed by de Sa~\citep{NIPS1993_831}, has seen a resurgence with the goal of learning strong feature representations that can be used for finetuning on downstream tasks.
Most works leverage audio-visual semantic correspondence~\citep{owens2016ambient, aytar2016soundnet, arandjelovic17look, m2020multimodal} or the synchronized timing of content~\citep{avts, owens2018audio} between the audio and visual streams.
Some works use this information to obtain within-clip sound localisation~\citep{owens2018audio,hu2019deep,arandjelovic2018objects,Senocak_2018,zhao2019sound,rouditchenko2019self} as well as audio-separation~\citep{gao2018learning,casanovas2010blind}.
Other methods use a modality distillation framework to learn video encoders from other modalities~\citep{owens2016ambient, piergiovanni2020evolving}. 
In~\citep{piergiovanni2020evolving}, a loss function is meta-learned by computing common self-supervised losses and distilling these and clustering is used as an evaluation metric for meta-learning.
New methods have started to learn even stronger representations by using ASR generated text from videos as another modality~\citep{sun2019videobert, sun2019contrastive, miech2019endtoend, li2020learning, nagrani2020}.

\paragraph{Clustering videos.}

Perhaps the simplest way of combining representation learning and clustering in videos is to apply \emph{post-hoc} a clustering algorithm after pretraining a representation.
In ClusterFit~\citep{yan2020cluster}, the authors show that running a simple $k$-means algorithm on the features from a pretrained network on the pretraining dataset yields small but consistent gains for representation learning when these clusters are used as labels and the networks are retrained.
While in~\citep{yan2020cluster}, the authors found the optimal number of clusters to consistently be at least one order of magnitude higher than the number of ground-truth labels, we investigate applying this method on various pretrained methods as baselines for our task of labelling an unlabelled video dataset.
Specifically, we apply $k$-means on state-of-the-art single modality models such as DPC~\citep{han2019video} and MIL-NCE~\cite{miech2019endtoend}, as well the multi-modal model XDC~\cite{alwassel2019self}, which itself uses $k$-means on the audio and visual streams to learn representations.
However, they do this as a pretext task for representation learning and obtain separate clusters for audio and video.
In contrast, our goal is multi-modally labelling an unlabelled dataset, and we find that our method works significantly better at this task.

\section{Method}

\begin{figure*}[t]
\centering
\includegraphics[width=0.95\textwidth]{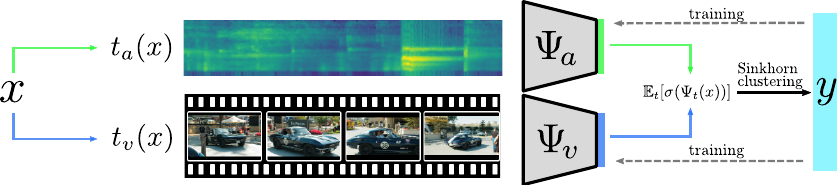}
\caption{\textbf{Our model} views modalities as different \textit{augmentations} and produces a multi-modal clustering of video datasets from scratch that can closely match human annotated labels.}
\label{fig:method}
\vspace{-10pt}
\end{figure*}

Given a dataset $D = \{\bx_i\}_{i \in \{1 ,\dots,N \}}$ of multi-modal data $\bx_i$, our goal is to learn a labelling function $y(\bx) \in \{1,\dots, K\}$ without access to any ground-truth label annotations.
There are two requirements that the labelling function must satisfy.
First, the labels should capture, as well as possible, the \emph{semantic content} of the data, in the sense of reproducing the labels that a human annotator would intuitively associate to the videos.
As part of this, we wish to account for the fact that semantic classes are not all equally probable, and tend instead to follow a Zipf distribution~\citep{youtube8m,kinetics}.
We then evaluate the quality of the discovered labels by matching them to the ones provided by human annotators, using datasets where ground-truth labels are known.
The second requirement is that the labelling method should not overly rely on a single modality.
Instead, we wish to treat each modality \textit{as equally informative} for clustering.
In this way, we can learn a more robust clustering function, which can work from either modality.
Furthermore, correlating of modalities has been shown to be a proxy to learn better abstractions~\cite{arandjelovic2018objects,avts,m2020multimodal,owens2018audio}.%

While our method can work with any number of data modalities (vision, audio, depth, textual transcripts, \dots), we illustrate it under the assumption of video data $\bx = (a,v)$, comprising an audio stream $a$ and a visual stream $v$.
The following two sections describe our method in detail and show how it meets our requirements.

\subsection{Non-degenerate clustering via optimal transport}

In this section, we briefly summarize the formulation of~\cite{asano2020self} to interpret clustering as an optimal transport problem.
SeLa~\cite{asano2020self} is a method that learns representations via clustering images.
The labelling function can be expressed as the composition $y(\Psi(\bx))$, where $\bz=\Psi(\bx)$ is a data representation (i.e.~a feature extractor implemented by a deep neural network), and $y(\bz) \in \{1,\dots,K\}$ operates on top of the features rather than the raw data.

Any traditional clustering algorithm, such as $k$-means or Gaussian mixture models, defines an energy function $E(y)$ that, minimized, gives the best data clustering function $y$.
When the representation is accounted for, the energy $E(y,\Psi)$ is a function of both $y$ and $\Psi$, and we may be na\"{\i}vely tempted to optimize over both.
However, this is well known to yield unbalanced  solutions, which necessitates ad-hoc techniques such as non-uniform sampling or re-initialization of unused clusters~\citep{caron2018deep,caron2019unsupervised}.
Theoretically, in fact, for most choices of $E$, the energy is trivially minimized by the representation $\Psi$ that maps all data to a constant.

\citet{asano2020self} address this issue by constraining the marginal probability distributions of the clusters  to be uniform, and show that this reduces to an optimal transport problem.
The algorithm then reduces to alternating the fast Sinkhorn-Knopp algorithm~\citep{sinkhornlightspeed} for clustering, and standard neural network training for representation learning.
To do this, one introduces the cross-entropy loss $E(q,p)$, between the labels given as one-hot vectors in $q$ (i.e.~$q(y(\bx)) =1~\forall \bx $) and the softmax outputs $p$ of a network $\Psi$:
\begin{equation}\label{e:cent}
E(p,q)
=
-
\frac{1}{N}
\sum_{i=1}^N\sum_{y=1}^K
q(y|\bx_i)
\log p(y|\bx_i),
~~~
p(y|\bx_i) = \operatorname{softmax} \Psi(\bx_i),
\end{equation}
where  $K$ is the number of clusters.
This energy is optimized under the constraint that the marginal cluster probability $\sum_{i=1}^{N} \frac{1}{N}  p(y|\bx_i) = \frac{1}{K}$ is constant (meaning all clusters are a-priori equally likely).
Note that minimizing $E$ with respect to $p$ is the same as training the deep network $\Psi$ using the standard cross-entropy loss.

Next, we show that minimizing $E(p,q)$ w.r.t.~the label assignments $q$ results in an optimal transport problem.
Let $P_{yi} =  p(y|\bx_i)\frac{1}{N}$ be the $K\times N$ matrix of joint probabilities estimated by the model and $Q_{yi} = q(y|\bx_i)\frac{1}{N}$ be $K\times N$ matrix of assigned joint probabilities.
Matrix $Q$ is relaxed to be an element of a transportation polytope
\begin{equation}\label{e:polytope}
U(r,c) \coloneqq
\{ Q \in \mathbb{R}^{K \times N}_+
~
\vert
~
Q \mathbbm{1} = r,
~
Q^\top \mathbbm{1} = c\}, \quad
r = \mathbbm{1}/K,
\quad
c = \mathbbm{1}/N.
\end{equation}
where $\mathbbm{1}$ are vectors of ones, and $r$ and $c$  the marginal projections of matrix $Q$ onto its clusters and data indices, respectively.
Finally, optimizing $E(P,Q)$ w.r.t.~to $Q \in U(r,c)$ is a linear optimal transport problem, for which~\citep{sinkhornlightspeed} provides a fast, matrix-vector multiplication based solution.

\subsection{Clustering with arbitrary prior distributions}

A shortcoming of the algorithm just described is the assumption that all clusters are equally probable.
This avoids converging to degenerate cases but is too constraining in practice since real datasets follow highly skewed distributions~\citep{youtube8m,kinetics}, and even in datasets that are collected to be uniform, they are not completely so~\citep{VGGSound,kinetics,tian2018ave}.
Furthermore, knowledge of the data distribution, for example long-tailedness, can be used as additional information (e.g. as in~\citep{piergiovanni2020evolving} for meta-learning) that can improve the clustering by allocating the right number of data points to each cluster.
Next, we describe a mechanism to change this distribution arbitrarily.

In the algorithm above, changing the label prior amounts to choosing a different cluster marginal $r$ in the polytope $U(r,c)$.
The difficulty is that $r$ is only known up to an arbitrary permutation of the clusters, as we do not know a-priori which clusters are more frequent and which ones less so.
To understand how this issue can be addressed, we need to explicitly write out the energy optimised by the Sinkhorn-Knopp (SK) algorithm~\citep{sinkhornlightspeed} to solve the optimal transport problem.
This energy is:
\begin{equation}
  \min_{Q \in U(r,c)} \langle Q, - \log P \rangle
  + 
  \frac{1}{\lambda} \operatorname{KL}(Q\|rc^\top),
\end{equation}
where $\lambda$ is a fixed parameter.
Let $r' = Rr$ where $R$ is a permutation matrix matching clusters to marginals.
We then seek to optimize the same quantity w.r.t.~$R$, obtaining the optimal permutation as $R^\ast = \operatorname{argmin}_{R} E(R)$ where
\begin{equation}
  E(R)
  = \langle Q, - \log P \rangle
  +
  \frac{1}{\lambda} \operatorname{KL}(Q\|Rrc^\top)
  =
  \text{const} + \sum_y - q(y)~[R \log r]_y. \label{eq:R}
\end{equation}
While there is a combinatorial number of permutation matrices, we show that minimizing \cref{eq:R} can be done by first sorting classes $y$ in order of increasing $q(y)$,
so that $y > y' \Rightarrow q(y) > q(y')$, and then finding the permutation that $R$ that also sorts $[R \log r]_y$ in increasing order.%
\footnote{
To see why this is optimal, and ignoring ties for simplicity, let $R$ be any permutation and construct a permutation $\bar R$ by applying $R$ and then by further swapping two labels $y>y'$. We can relate the energy of $R$ and $\bar R$ as:
\begin{equation}
\begin{aligned}
 E(R)
 &=
 E(\bar R)
 + q(y)[\bar R\log r]_{y} + q(y') [\bar R\log r]_{y'}
 - q(y)[\bar R\log r]_{y'} - q(y') [\bar R\log r]_{y}
 \\
 &=
 E(\bar R) + (q(y) - q(y'))~([\bar R\log r]_{y} - [\bar R\log r]_{y'}).
\end{aligned}
\end{equation}
Since the first factor is positive by assumption, this equation shows that the modified permutation $\bar R$ has a lower energy than $R$ if, and only if, $[\bar R\log r]_{y} > [\bar R\log r]_{y'}$, which means that $\bar R$ sorts the pair in increasing order.
}
We conclude that $R$ cannot be optimal unless it sorts all pairs.
After this step, the SK algorithm can be applied using the optimal permutation $R^*$, without any significant cost (as solving for $R$ is equivalent to sorting $\mathcal{O}(K\log K)$ with $K \ll N$).
The advantage is that it allows to choose any marginal distribution, even highly unbalanced ones which are likely to be a better match for real world image and video classes than a uniform distribution.

\subsection{Multi-modal single labelling}

Next, we tackle our second requirement of extracting as much information as possible from multi-modal data.
In principle, all we require to use the clustering formulation \cref{e:cent} with multi-modal data $\bx = (a,v)$ is to design a corresponding multi-modal representation $\Psi(\bx)=\Psi(a,v)$.
However, we argue for \textit{multi-modal single labelling} instead.
By this, we mean that we wish to cluster data one modality at a time, but in a way that is modality agnostic.
Formally, we introduce \emph{modality splicing transformations}~\citep{m2020multimodal} $t_a(\bx) =a$ and $t_v(\bx)=v$ and use these as \emph{data augmentations}.
Recall that augmentations are random transformations $t$ such as rotating an image or distorting an audio track that one believes should leave the label/cluster invariant.
We thus require our activations used for clustering to be an average over augmentations by replacing matrix $\log P$ with
\begin{equation}
  \label{e:modality_avg}
  [\log P]_{yi} = \mathbb{E}_t[\log \operatorname{softmax}_y  \Psi(t\bx_i)].
\end{equation}
If we consider splicing as part of the augmentations, we can learn clusters that are invariant to standard augmentations \emph{as well as} the choice of modality.
In practice, to account for modality splicing, we define and learn a pair $\Psi=(\Psi_a,\Psi_v)$ of representations, one per modality, resulting in the same clusters ($\Psi_a(t_a(\bx)) \approx \Psi_v(t_v(\bx))$).
This is illustrated in \Cref{fig:method}.

\paragraph{Initialization and alignment.}

Since networks $\Psi_a$ and $\Psi_v$ are randomly initialized, at the beginning of training their output layers are \emph{de-synchronized}.
This means that there is no reason to believe that $\Psi_a(t_a(\bx))) \approx \Psi_v(t_v(\bx))$ simply because the \emph{order} of the labels in the two networks is arbitrary.
Nevertheless, in many self-supervised learning formulations, one exploits the fact that even randomly initialized networks capture a useful data prior~\cite{ulyanov17deepimageprior}, which is useful to bootstrap learning.

In order to enjoy a similar benefit in our formulation, we propose to \emph{synchronise} the two output layers of $\Psi_a$ and $\Psi_b$ before training the model.
Formally, we wish to find the permutation matrix $R$ that, applied to the last layer of one of the two encoders maximizes the agreement with the other (still leaving all the weights to their initial random values).
For this, let $W_a$ and  $W_v$ be the last layer weight matrices of the two networks,\footnote{We assume that the linear layer biases are incorporated in the weight matrices.} such as
$\Psi_a(a) = W_a \bar \Psi_a(a)$ and
$\Psi_v(v) = W_v \bar \Psi_v(v)$.
We find the optimal permutation $R$ by solving the optimisation problem:
\begin{equation}
\min_R
\sum_{i=1}^{N}
\left\vert
\operatorname{softmax}(
  RW_a \bar \Psi_a(t_a(\bx_i))
)
-
\operatorname{softmax}(
  W_v \bar \Psi_v(t_v(\bx_i))
) \right\vert, \label{eq:av-align}
\end{equation}
In order to compare softmax distributions, we choose $\vert \cdot \vert$ as the 1-norm, similar to~\citep{hinton2015distilling}.
We optimize \cref{eq:av-align} with a greedy algorithm: starting with a feasible solution and switching random pairs when they reduce the cost function~\citep{christofides1976worst}, as these are quick to compute and we do not require the global minimum. 
Further details are given in~\cref{appx:pairwise-details}. 
With this permutation, the weight matrix of the last layer of one network can be resorted to match the other.

\paragraph{Decorrelated clustering heads.}

Conceptually, there is no single `correct' way of clustering a dataset: for example, we may cluster videos of animals by their species, or whether they are taken indoor or outdoor.
In order to alleviate this potential issue, inspired by~\citep{asano2020self,ji2018invariant}, we simply learn multiple labelling functions $y$, using multiple classification heads for the network.
We improve this scheme as follows.
In each round of clustering, we generate two random augmentations of the data.
Then, the applications of SK to half of the heads (at random) see the first version, and the other half the second version, thus increasing the variance of the resulting clusters.
This increases the cost of the algorithm by only a small amount --- as more time is used for training instead of clustering.

\section{Experiments}\label{sec:experiments}

The experiments are divided into three parts.
First, in \cref{sec:analysis}, we analyze the need for using both modalities when clustering and investigate the effect of our individual technical contributions via ablations and comparison to other approaches.
Second, in \cref{sec:labelling}, we demonstrate how our method achieves its stated goal of labelling a video dataset without human supervision.
Third, in \cref{sec:repr-learning}, we show that a side effect of our method is to learn effective audio-visual representations that can be used for downstream tasks \eg video action retrieval,  establishing a new state of the art.

\paragraph{Datasets.}

While the goal and target application of this work is to group unlabelled video datasets, for analysis purposes only, we use datasets that contain human annotated labels.
The datasets range from small- to large-scale:
The first is the recently released \textbf{VGG-Sound}~\citep{VGGSound}, which contains around 200k videos obtained in the wild from YouTube with low labelling noise and covering 309 categories of general classes.
The second dataset is \textbf{Kinetics-400}~\citep{kinetics}, which contains around 230k videos covering 400 human action categories.
Third, we test our results on \textbf{Kinetics-Sound} proposed in~\citep{arandjelovic17look}, formed by filtering the Kinetics dataset to 34 classes that are potentially manifested visually and audibly, leading to 22k videos. %
Lastly, we use the small-scale \textbf{AVE Dataset}~\citep{tian2018ave}, originally proposed for audio-visual event localization and containing only around 4k videos.
Among these, only VGG-Sound and Kinetics-400 are large enough for learning strong representations from scratch. 
We therefore train on these datasets and unsupervisedly finetune the VGG-Sound model on Kinetics-Sound and AVE.

\paragraph{Training details.}
Our visual encoder is a R(2+1)D-18~\citep{Tran18} network and our audio encoder is a ResNet~\citep{KaimingHe16} with 9 layers.
For optimization, we use SGD for 200 epochs with weight decay of $10^{-5}$ and momentum of $0.9$, further implementation details are provided in~\cref{appx:training-details}.

\paragraph{Baselines.}

\begin{table}[!htb]
    \centering
    \footnotesize
    \caption{\textbf{Architectures and pretraining datasets.} We use state-of-the-art representation learning methods and combine pretrained representations with $k$-means as baselines in the  \cref{tab:vggs,tab:ave,tab:k400,tab:k-sound}. \label{tab:details-visual}}
    \vspace{1em}
    \begin{tabular}{l r l l}
    \toprule
        Method & Input shape                       & Architecture & Pretrain dataset \\
         \midrule
        Supervised                          & $32\times3\times112\times112$ & R(2+1)D-18   & Kinetics-400 \\ 
        DPC~\citep{han2019video}            & $40\times3\times224\times224$ & R3D-34       & Kinetics-400 \\
        MIL-NCE~\cite{miech2019endtoend}    & $32\times3\times224\times224$ & S3D          & HowTo100M\\
        XDC~\cite{alwassel2019self}         & $32\times3\times224\times224$ & R(2+1)D-18   & Kinetics-400\\
    \bottomrule
    \end{tabular}
\end{table}
To compare our method on this novel task of clustering these datasets, we obtained various pretrained video representations (DPC~\citep{han2019video}, MIL-NCE~\cite{miech2019endtoend}and XDC~\cite{alwassel2019self}), both supervised%
\footnote{
The R(2+1)D-18 model from PyTorch trained on Kinetics-400~\citep{kinetics} from \url{https://github.com/pytorch/vision/blob/master/torchvision/models/video/resnet.py}.
}
and self-supervised (see \cref{tab:details-visual} for details).
For comparison, following~\citep{yan2020cluster}, we run $k$-means on the global-average-pooled features, setting $k$ to the same number of clusters as our method to ensure a fair comparison.
For the $k$-means algorithm, we use the GPU accelerated version from the Faiss library~\citep{faiss}.

\paragraph{Evaluation.} We adopt standard metrics from the self-supervised and unsupervised learning literature: the \emph{normalized mutual information} (NMI), the \emph{adjusted rand index} (ARI) and \emph{accuracy} (Acc)  after matching of the self-supervised labels to the ground truth ones (for this we use the Kuhn–Munkres/Hungarian algorithm~\citep{kuhn55hungarian}).
We also report the \emph{mean entropy} and the \emph{mean maximal purity} per cluster, defined in~\cref{appx:entropy-purity}, to analyze the qualities of the clusters.
For comparability and interpretability,  we evaluate the results using the ground truth number of clusters -- which usually is unknown -- but we find our results to be stable w.r.t.~other number of clusters (see Appendix).

\subsection{Technical Analysis \label{sec:analysis}}

\paragraph{Multi-modality.}
In order to shed light on the nature of labelling a multi-modal dataset, we provide a detailed study of the use and combination of modalities in \cref{tab:ablation}.
While visual-only methods such as the Kinetics-400 supervisedly pretrained model, MIL-NCE or DPC cannot work when only audio-stream ({\setlength{\fboxsep}{1.2pt}\colorbox{blue!30}{\faVolumeUp}}) is provided, we show the results for XDC and our method when a single or both modalities are provided.
In particular, we find that even when only the visual-stream ({\setlength{\fboxsep}{1.7pt}\colorbox{green!30}{\faVideoCamera}}) is present at test-time, our method (57\% NMI) already outperforms methods solely developed for representation learning, even surpassing the 100M videos with transcripts trained MIL-NCE (49\% NMI).
When only the audio-stream is used, our method's performance drops only slightly, congruent with the balanced informativeness of both modalities in our method.
Finally, when both modalities are used, we find that our method profits from both, such that the combined performance is significantly higher than the maximal performance of each single modality alone.

\paragraph{Degraded modality.} 
In \cref{fig:compression}, we analyze how well our method fares when the quality of one modality is reduced. 
For this, we compress the video-stream by down and upsampling of the video resolution by factors 1, 4, 8 and 16 (details are provided in~\cref{degraded-detail}).
Even though our method has not been trained with compressed videos, we find that its performance degrades more gracefully than the baselines indicating it has learned to rely on both modalities.

\paragraph{Ablation.}
\begin{table}[tb]
\footnotesize
\begin{minipage}[b]{.60\textwidth}
\setlength{\tabcolsep}{3pt}
   \centering
   \caption{\textbf{The value of multi-modal understanding} is observed for obtaining a strong set of labels for VGG-Sound.
  Our method combines both modalities effectively to yield accurracies beyond a single modality and other methods. \label{tab:ablation}}
  \vspace{4pt}
  \setlength{\fboxsep}{1.2pt}
  \begin{tabular}{l @{\hskip 5pt}c c c c c c c}
  \toprule
     Method             & \colorbox{blue!30}{\faVolumeUp} & {\setlength{\fboxsep}{1.7pt}\colorbox{green!30}{\faVideoCamera}}  &   \textbf{NMI} & \textbf{ARI} & \textbf{Acc.} & $\langle \mathbf{H} \rangle$ & $\langle \mathbf{p_\mathrm{max}} \rangle$ \\
    \midrule
    Random              & \xmark & \cmark & 10.2 & 4.0 & 2.2 & 4.9 & 3.5 \\
    Supervised          & \xmark & \cmark & 46.5 & 15.6 & 24.3 & 2.9 & 30.8 \\
    \midrule
    DPC                 & \xmark & \cmark & 15.4 & 0.7 & 3.2 & 4.7 & 4.9 \\
    MIL-NCE             & \xmark & \cmark & 48.5 & 12.5 & 22.0 & 2.6 & 32.9 \\
    \midrule
    XDC                 & \xmark & \cmark & 16.7 & 1.0 & 3.9 & 4.5 & 6.4 \\
                        & \cmark & \xmark & 14.0 & 0.8 & 2.9 & 4.6 & 4.4 \\
                        & \cmark & \cmark & 18.1 & 1.2 & 4.5 & 4.41 & 7.4 \\
    \midrule
    \textbf{SeLaVi}       & \xmark & \cmark & 52.8 & 19.7 & 30.1 & 2.6 & 35.6 \\
                        & \cmark & \xmark & 47.5 & 15.2 & 26.5 & 2.8 & 32.9 \\
                        & \cmark & \cmark &  \bf{55.9} & \bf{21.6} & \bf{31.0} & \bf{2.5} & \bf{36.3} \\
    \bottomrule
  \end{tabular}
  \vspace{8pt}
\end{minipage}
\hfill
\begin{minipage}[b]{.37\textwidth}
\centering
    \includegraphics[width=0.95\textwidth]{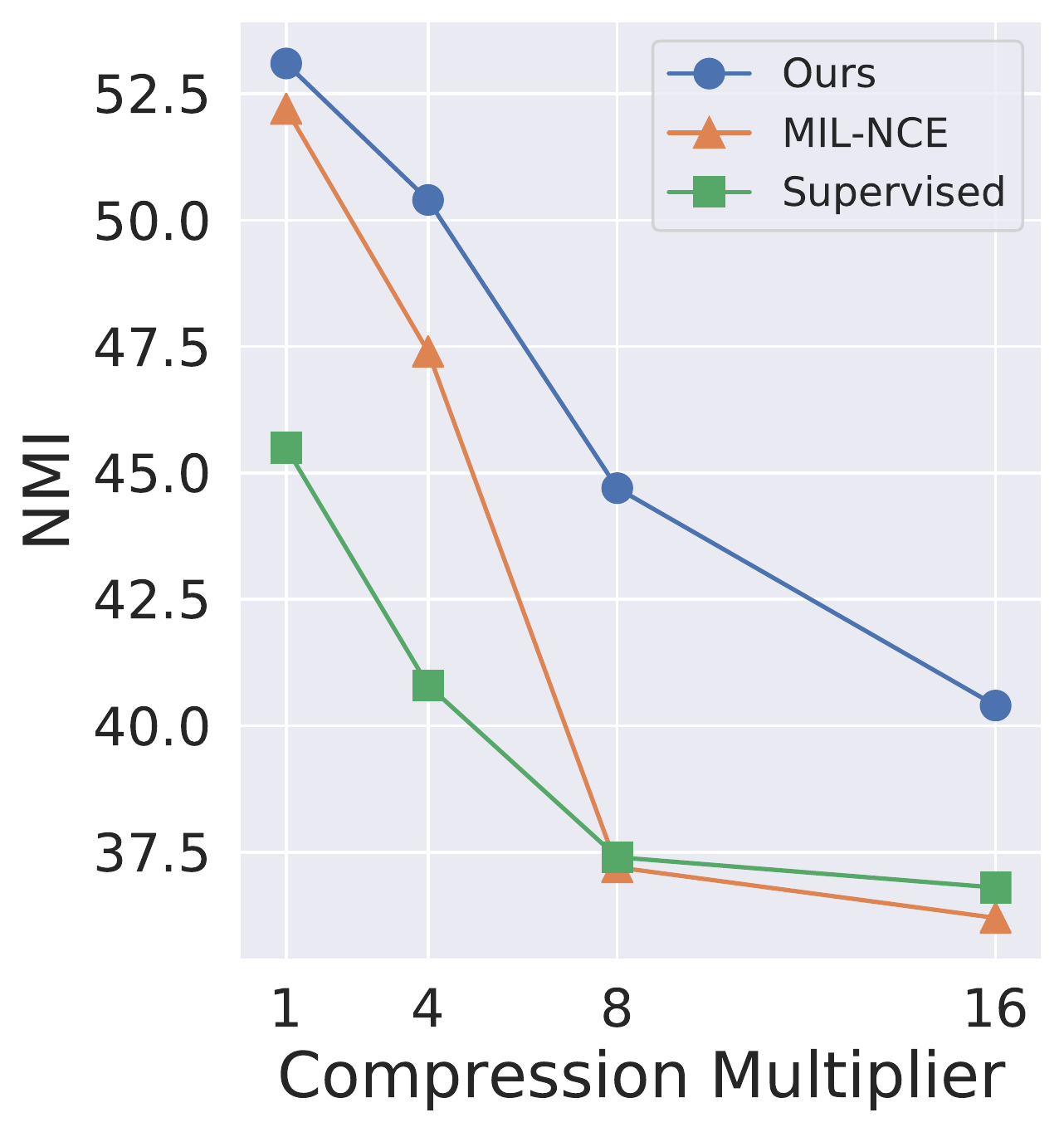}
    \vspace{2mm}
    \captionof{figure}{\textbf{Effective use of multi-modality} is found for our method when the visual input is compressed and decompressed. \label{fig:compression}
    }
\end{minipage}
\end{table}
\begin{table}
\vspace{-2em}
\footnotesize
\begin{minipage}[t]{.49\textwidth}

\footnotesize
\setlength{\tabcolsep}{1.5pt}
  \setlength{\fboxsep}{1.2pt}
  \centering
  \captionof{table}{ \textbf{Ablation}
  of multi-modality, \ul{M}odality \ul{A}lignment and
  \ul{G}aussian  marginals. 
  \ul{D}ecorrelated \ul{H}eads.  Models are evaluated at 75 epochs on the VGG-Sound dataset.
  \label{tab:ablation-ours}}
  \begin{tabular}{l @{\hskip 5pt} cccccc @{\hskip 5pt} c c }
  \toprule
     Method         &  \faFileImageO &  (\colorbox{blue!30}{\faVolumeUp} {\setlength{\fboxsep}{1.7pt}\colorbox{green!30}{\faVideoCamera}}) & MA? & G.? & DH?  & \textbf{Acc} & \textbf{ARI} & \textbf{NMI}\\ %
    \midrule                                                         %
     (a) SeLa            & \cmark & \xmark & --      & --     & --      & 6.4   & 2.3   & 20.6     \\ 
     (b) Concat          & \xmark & \cmark & --      & \xmark & \xmark  & 7.6 & 3.2 & 24.7        \\
     (c) SeLaVi          & \xmark & \cmark & \xmark  & \xmark & \xmark  & 24.6 & 15.6 & 48.8      \\ 
    \midrule
     (d) SeLaVi          & \xmark & \cmark & \xmark  & \cmark & \cmark  & 26.6 & 18.5  & 50.9      \\
     (e) SeLaVi          & \xmark & \cmark & \cmark  & \xmark & \cmark   & 26.2 & 17.3  & 51.5     \\
     (f) SeLaVi          & \xmark & \cmark & \cmark  & \cmark & \xmark  & 23.9 & 14.7  & 49.9       \\
    \midrule
     (g) \textbf{SeLaVi} & \xmark & \cmark & \cmark & \cmark  & \cmark  & 26.6 & 17.7  & 51.1 \\
     \bottomrule
  \end{tabular}
  
\end{minipage}
\hfill
\begin{minipage}[t]{.45\textwidth}
\setlength{\tabcolsep}{2.2pt}
\centering
  \footnotesize
    \caption{\textbf{Retrieval}\label{tab:ret} via various number of nearest neighbors.\label{tab:retrieval}}
       \vspace{4pt}
  \begin{tabular}{l c c c @{\hskip 5pt} c c c}
  \toprule
       & \multicolumn{3}{c}{\textbf{HMDB}} & \multicolumn{3}{c}{\textbf{UCF}} \\
       \cmidrule(lr){2-4} \cmidrule(lr){5-7} \multicolumn{1}{r}{Recall @} & 1 & 5 & 20 & 1 & 5  & 20    \\
    \midrule %
    3D-Puzzle~\citep{kim2019self}             & --  & --   & --   & 19.7 & 28.5  & 40.0 \\
    OPN~\citep{lee2017unsupervised}           & --  & --   & --   & 19.9 & 28.7  & 40.6   \\
    ST Order ~\citep{buchler2018improving}    & --  & --   & --   & 25.7 & 36.2  & 49.2  \\
    ClipOrder~\citep{clip_order}              & 7.6 & 22.9 & 48.8 & 14.1 & 30.3  & 51.1   \\
    SpeedNet~\citep{benaim2020speednet}       & --  & --   & --   & 13.0 & 28.1  & 49.5  \\
    VCP~\citep{luo2020video}                  & 7.6 & 24.4 & 53.6 & 18.6 & 33.6  & 53.5   \\
    VSP~\citep{cho2020selfsupervised}         & 10.3& 26.6 & 54.6 & 24.6 & 41.9  & 76.9   \\
    \midrule
      \textbf{SeLaVi} & \bf{24.8} & \bf{47.6}   & \bf{75.5} &  \bf{52.0} & \bf{68.6}  & \bf{84.5}     \\
    \bottomrule
  \end{tabular}
 
\end{minipage}
\end{table}
Next, we ablate the key parameters of our method and show how they each contribute to the overall clustering quality in~\cref{tab:ablation-ours}.
First, we show a baseline model in \cref{tab:ablation-ours}(a), when na\"ively applying the publically available source-code for SeLa~\citep{asano2020self} on video frames (\faFileImageO), this yields a NMI of 20\%.
Compared to this frame-only method, in row (b), we find the results for  concatenating the features of both modalities (followed by a single clustering head) to only lead to a small improvement upon the frame-only method with a NMI of $25\%$. %
Our method is shown in row (c), where we find a substantial improvement with a NMI of $52\%$, i.e. a relative gain of more than 100\%.
While part of the gain comes from multi-modality, especially compared to row (b), the largest gain comes from the ability of our method in exploiting the natural correspondence provided in the multi-modal data. 
Finally, by ablating the technical improvements in rows (d)-(f) we find the strongest gain to be coming decorrelated heads, followed by the audio-visual modality alignment (MA) procedure, and that each improvement indeed benefits the model. 
To analyze the gains obtained by using multiple heads, we have also computed the average NMI between all pairs of heads as $(77.8\pm 4\%)$. 
This means that the different heads do learn fairly different clusterings (as NMI takes permutations into account) whilst being at a similar distance to the `ground-truth' ($53.1\pm 0.1\%)$.

\subsection{Unsupervised labelling audio-visual data \label{sec:labelling}}

\begin{table*}[tb]
\centering
\caption{\textbf{Unsupervised labelling of datasets.} 
	We compare labels from our method to labels that are obtained with $k$-means on the representations from a supervised and various unsupervised methods on four datasets. 	}\label{tab:labelling-datasets}
	\vspace{-1mm}
	\begin{subtable}[t]{0.45\linewidth}\centering
		{\setlength{\tabcolsep}{3pt}
   \centering
   \footnotesize
   \caption{\textbf{VGG-Sound.}\label{tab:vggs}}
   \vspace{-4pt}
  \begin{tabular}{l @{\hskip 5pt} c c c c c}
  \toprule
     Method                        &   \textbf{NMI} & \textbf{ARI} & \textbf{Acc.} & $\langle \mathbf{H} \rangle$ & $\langle \mathbf{p_\mathrm{max}} \rangle$ \\
    \midrule 
    Random        & 10.2 & 4.0 & 2.2 & 4.9 & 3.5 \\
    Supervised    & 46.5 & 15.6 & 24.3 & 2.9 & 30.8 \\ 
    \midrule
    DPC           & 15.4 & 0.7 & 3.2 & 4.7 & 4.9 \\
    XDC           & 18.1 & 1.2 & 4.5 & 4.41 & 7.4 \\
    MIL-NCE       & 48.5 & 12.5 & 22.0 & 2.6 & 32.9 \\
    \midrule    
    \textbf{SeLaVi} &  \bf{55.9} & \bf{21.6} & \bf{31.0} & \bf{2.5} & \bf{36.3} \\
    \bottomrule
  \end{tabular}
  \vspace{7pt}
}%
	\end{subtable}
	\qquad 
	\begin{subtable}[t]{0.45\linewidth}\centering
		{\setlength{\tabcolsep}{3pt}
   \centering
   \footnotesize
\caption{\textbf{AVE.}\label{tab:ave}}
\vspace{-4pt}
  \begin{tabular}{l @{\hskip 5pt} c c c c c}
  \toprule
     Method                        &   \textbf{NMI} & \textbf{ARI} & \textbf{Acc.} & $\langle \mathbf{H} \rangle$ & $\langle \mathbf{p_\mathrm{max}} \rangle$ \\
    \midrule 
    Random     &  9.2 & 1.3 & 9.3 & 2.9 & 12.6 \\
    Supervised & 58.4 & 34.8 & 50.5 & 1.1 & 60.6\\ 
    \midrule
    DPC        & 18.4 & 5.0 & 15.1 & 2.7 & 17.5 \\
    XDC        & 17.1 & 6.0 & 16.4 & 2.6 & 19.1 \\
    MIL-NCE    & 56.3 & 30.3 & 42.6 & \bf{1.2} & 57.1\\
    \midrule
    \textbf{SeLaVi} & \textbf{66.2} & \bf{47.4} & \textbf{57.9} & \textbf{1.1} & \textbf{59.3}  \\
    \bottomrule
  \end{tabular}
  \vspace{7pt}
  
}	
	\end{subtable} \\
	\begin{subtable}[t]{0.45\linewidth}\centering
		{\setlength{\tabcolsep}{3pt}
   \centering
   \footnotesize
   \caption{\textbf{Kinetics.}\label{tab:k400}}
   \vspace{-4pt}
  \begin{tabular}{l @{\hskip 5pt} c c c c c}
  \toprule
     Method                        &   \textbf{NMI} & \textbf{ARI} & \textbf{Acc.} & $\langle \mathbf{H} \rangle$ & $\langle \mathbf{p_\mathrm{max}} \rangle$ \\
    \midrule 
    Random & 11.1 & 0.2 & 1.8 & 5.1 & 3.3 \\
    Supervised & 70.5 & 43.4 & 54.9 & 1.6 & 62.2 \\
    \midrule
    DPC & 16.1 & 0.6 & 2.7 & 4.9 & 3.9 \\
    XDC & 17.2 & 0.8 & 3.4 & 4.7 & 6.2 \\
    MIL-NCE & \textbf{48.9} & \textbf{12.5} & \textbf{23.5} & \textbf{2.7} & \textbf{33.7} \\
    \midrule
    \textbf{SeLaVi} & 27.1 & 3.4 & 7.8 & 4.8 & 9.4 \\
    \bottomrule
  \end{tabular}
  
  \vspace{2pt}
}%
	\end{subtable}
	\qquad 
	\begin{subtable}[t]{0.45\linewidth}\centering
		{\setlength{\tabcolsep}{3pt}
   \centering
   \footnotesize
   \caption{\textbf{Kinetics-Sound.}\label{tab:k-sound}}
   \vspace{-4pt}
  \begin{tabular}{l @{\hskip 5pt} c c c c c}
  \toprule
     Method                        &   \textbf{NMI} & \textbf{ARI} & \textbf{Acc.} & $\langle \mathbf{H} \rangle$ & $\langle \mathbf{p_\mathrm{max}} \rangle$ \\
    \midrule 
    Random        & 2.8        & 0.5      & 5.9       & 3.3      & 8.3 \\
    Supervised    & 81.7       & 66.3     & 75.0      & 0.5      & 85.4 \\ 
    \midrule            
    DPC           & 8.8       & 2.2       & 9.6       & 3.1      & 13.6 \\
    XDC           & 7.5       & 1.9       & 9.4       & 3.1      & 13.6 \\
    MIL-NCE       & \bf{47.5}       & 24.0      & 37.8      & \textbf{1.5} & \bf{51.0} \\

    \midrule
    \textbf{SeLaVi} & \bf{47.5} & \bf{28.7} & \bf{41.2} & 1.8      & 45.5 \\
    \bottomrule
  \end{tabular}
 
  \vspace{2pt}
  
}	
	\end{subtable}
\vspace{-10pt}
\end{table*}
\Cref{tab:labelling-datasets} shows the quality of the labels obtained automatically by our algorithm.
We find that for the datasets VGG-Sound, Kinetics-Sound, and AVE, our method achieves state-of-the-art clustering performance with high accuracies of $55.9\%, 41.2\%, 57.9\%$, even surpassing the one of the strongest video feature encoder at present, the manually-supervised R(2+1)D-18 network.
This result echoes the findings in the image domain~\citep{gansbeke2020learning} where plain $k$-means on representations is found to be less effective compared to learning clusters.
For Kinetics-400, we find that the clusters obtained from our method are not well aligned to the human annotated labels. %
This difference can explained by the fact that Kinetics is strongly focused on visual (human) actions and thus the audio is given almost no weighting in the annotation.
We encourage exploring our interactive material, where our method finds clusters grouped by similar background music, wind or screaming crowds. 
We stress that such a grouping is \textit{ipso facto} not wrong, only not aligned to this set of ground truth labels.

\subsection{Labelling helps representation learning \label{sec:repr-learning}}
Finally, we show how the visual feature representations unsupervisedly obtained from our method perform on downstream tasks.
While not the goal of this paper, we test our representation on a standardized video action retrieval task in \cref{tab:retrieval} and also provide results on video action classification in \cref{tab:visual-downstream}, and refer to the Appendix for implementation details.
We find that in obtaining strong labels, our method simultaneously learns robust, visual representations that can be used for other tasks without any finetuning and significantly improve the state of the art by over 100\% for Recall @1 on UCF-101 and HMDB-51.
\section{Conclusion}
In this work, we have
established strong baselines for the problem of unsupervised labelling of several popular video datasets;
introduced a simultaneous clustering and representation learning approach for multi-modal data that outperforms all other methods on these benchmarks; and
analysed the importance of multi-modality for this task in detail.
We have further found that strong representations are not a sufficient criterion for obtaining good clustering results, yet, the strongest feature representations remain those obtained by supervised, \ie~well-clustered training.
We thus expect the field of multi-modal clustering to be rapidly adopted by the research community who can build upon the presented method and baselines. 

\section*{Broader Impact}

We propose a method for clustering videos automatically.
As such, we see two main areas of potential broader impact on the community and society as a whole.

\paragraph{Few-label harmful content detection.}
Our method clusters a video dataset into multiple sets of similar videos, as evidenced by the audio- and visual-stream and produces consistent, homogenous groupings.
In practice, unsupervised clustering is especially useful for reducing the amount of data that human annotators have to label, since for highly consistent clusters only a single label needs to be manually obtained which can be propagated to the rest of the videos in the cluster.
Using such an approach for the purpose of detecting harmful online content is especially promising. 
In addition, label-propagation might further lead to a beneficial reduction of type I errors (saying a video is safe when it is not).
Furthermore, the multi-modality of our method allows it to potentially detect harmful content that is only manifested in one modality such as static background videos of harmful audio.
Multi-modal harmful content detection has also been a subject of a recent data challenge that emphasizes insufficiency of using a single modality\footnote{Hateful memes challenge: \url{https://www.drivendata.org/competitions/64/hateful-memes/}}.
Lastly, the generality of our method allows it to also scale beyond these two modalities and in the future also include textual transcripts.
Given the importance of this topic, it is also important to acknowledge, while less of a direct consequence, potential biases that can be carried by the dataset.
Indeed models trained using our method will inherit the biases present in the dataset, which could be known but also unknown, potentially leading to propagation of biases without a clear way to analyze them, such as via labels.
However, given the numerous pitfalls and failures when deploying computer vision systems to the real world, we believe that the positive impact of foundational research on public datasets, such as is presented in this paper, far outweighs these risks lying further downstream.

\paragraph{Overestimating clustering quality.}
The main benefit of our approach is to reduce the cost of grouping large collections of video data in a `meaningful' way.
It is difficult to think of an application where such a capability would lead directly to misuse.
In part, this is due to the fact that better clustering results can generally be obtained by using some manual labels, so even where clustering videos could be misused, this probably would not be the method of choice.
Perhaps the most direct risk is that a user of the algorithm might overestimate its capabilities.
Clustering images is sometimes done in critical applications (e.g.~medical science \citep{jiang2019novel,li2019clu,iakovidis2018detecting}).
Our method clusters data based on basic statistical properties and the inductive prior of convolutional neural networks, without being able to tap into the deep understanding that a human expert would have of such domain expertise.
Hence, the clusters determined by our method may not necessarily match the clusters an expert would make in a particular domain. 
Further, as the method is unsupervised, it may learn to exploit biases present in the data that might not be desired by the users.
While we believe it has potential to be broadly applied after being finetuned to a specific domain, at present, our method is a better fit for applications such as indexing personal video collections where clustering `errors' can be tolerated.

\begin{ack}
We are grateful for support from the Rhodes Trust (M.P.), Qualcomm Innovation Fellowship (Y.A.) and EPSRC Centre for Doctoral Training in Autonomous Intelligent Machines \& Systems [EP/L015897/1] (M.P. and Y.A.). M.P. funding was received under his Oxford affiliation. C.R. is supported by ERC IDIU-638009.
We thank Weidi Xie and Xu Ji from VGG for fruitful discussions.
\end{ack}

\paragraph{Erratum}
In our initial version there was a bug in our code and we have since updated our repository and updated results in Tables 2,3 and 5 in this paper.

\setlength{\bibsep}{0pt plus 0.3ex}
{\small\bibliographystyle{plainnat}\bibliography{shortstrings,refs}}

\begin{thebibliography}{82}
\providecommand{\natexlab}[1]{#1}
\providecommand{\url}[1]{\texttt{#1}}
\expandafter\ifx\csname urlstyle\endcsname\relax
  \providecommand{\doi}[1]{doi: #1}\else
  \providecommand{\doi}{doi: \begingroup \urlstyle{rm}\Url}\fi

\bibitem[Abu-El-Haija et~al.(2016)Abu-El-Haija, Kothari, Lee, Natsev, Toderici,
  Varadarajan, and Vijayanarasimhan]{youtube8m}
Sami Abu-El-Haija, Nisarg Kothari, Joonseok Lee, Paul Natsev, George Toderici,
  Balakrishnan Varadarajan, and Sudheendra Vijayanarasimhan.
\newblock {YouTube-8M}: A large-scale video classification benchmark.
\newblock \emph{arXiv preprint arXiv:1609.08675}, 2016.

\bibitem[Alwassel et~al.(2019)Alwassel, Mahajan, Torresani, Ghanem, and
  Tran]{alwassel2019self}
Humam Alwassel, Dhruv Mahajan, Lorenzo Torresani, Bernard Ghanem, and Du~Tran.
\newblock Self-supervised learning by cross-modal audio-video clustering.
\newblock \emph{arXiv preprint arXiv:1911.12667}, 2019.

\bibitem[Arandjelovic and Zisserman(2017)]{arandjelovic17look}
Relja Arandjelovic and Andrew Zisserman.
\newblock Look, listen and learn.
\newblock In \emph{Proc. ICCV}, 2017.

\bibitem[Arandjelovi\'{c} and Zisserman(2018)]{arandjelovic2018objects}
Relja Arandjelovi\'{c} and Andrew Zisserman.
\newblock Objects that sound.
\newblock In \emph{ECCV}, 2018.

\bibitem[Asano et~al.(2020{\natexlab{a}})Asano, Rupprecht, and
  Vedaldi]{asano2020a}
Yuki~M. Asano, Christian Rupprecht, and Andrea Vedaldi.
\newblock A critical analysis of self-supervision, or what we can learn from a
  single image.
\newblock In \emph{ICLR}, 2020{\natexlab{a}}.

\bibitem[Asano et~al.(2020{\natexlab{b}})Asano, Rupprecht, and
  Vedaldi]{asano2020self}
Yuki~M Asano, Christian Rupprecht, and Andrea Vedaldi.
\newblock Self-labelling via simultaneous clustering and representation
  learning.
\newblock In \emph{ICLR}, 2020{\natexlab{b}}.

\bibitem[Aytar et~al.(2016)Aytar, Vondrick, and Torralba]{aytar2016soundnet}
Yusuf Aytar, Carl Vondrick, and Antonio Torralba.
\newblock Soundnet: Learning sound representations from unlabeled video.
\newblock In \emph{NeurIPS}, 2016.

\bibitem[Bachman et~al.(2019)Bachman, Hjelm, and
  Buchwalter]{bachman2019learning}
Philip Bachman, R~Devon Hjelm, and William Buchwalter.
\newblock Learning representations by maximizing mutual information across
  views, 2019.

\bibitem[Bautista et~al.(2016)Bautista, Sanakoyeu, Tikhoncheva, and
  Ommer]{bautista2016cliquecnn}
Miguel~A Bautista, Artsiom Sanakoyeu, Ekaterina Tikhoncheva, and Bjorn Ommer.
\newblock Cliquecnn: Deep unsupervised exemplar learning.
\newblock In \emph{NeurIPS}, pages 3846--3854, 2016.

\bibitem[Bautista et~al.(2017)Bautista, Sanakoyeu, and Ommer]{bautista2017deep}
Miguel~A Bautista, Artsiom Sanakoyeu, and Bjorn Ommer.
\newblock Deep unsupervised similarity learning using partially ordered sets.
\newblock In \emph{CVPR}, pages 7130--7139, 2017.

\bibitem[Benaim et~al.(2020)Benaim, Ephrat, Lang, Mosseri, Freeman, Rubinstein,
  Irani, and Dekel]{benaim2020speednet}
Sagie Benaim, Ariel Ephrat, Oran Lang, Inbar Mosseri, William~T. Freeman,
  Michael Rubinstein, Michal Irani, and Tali Dekel.
\newblock Speednet: Learning the speediness in videos, 2020.

\bibitem[Buchler et~al.(2018)Buchler, Brattoli, and
  Ommer]{buchler2018improving}
Uta Buchler, Biagio Brattoli, and Bjorn Ommer.
\newblock Improving spatiotemporal self-supervision by deep reinforcement
  learning.
\newblock In \emph{Proceedings of the European Conference on Computer Vision
  (ECCV)}, pages 770--786, 2018.

\bibitem[Caron et~al.(2018)Caron, Bojanowski, Joulin, and Douze]{caron2018deep}
Mathilde Caron, Piotr Bojanowski, Armand Joulin, and Matthijs Douze.
\newblock Deep clustering for unsupervised learning of visual features.
\newblock In \emph{ECCV}, 2018.

\bibitem[Caron et~al.(2019)Caron, Bojanowski, Mairal, and
  Joulin]{caron2019unsupervised}
Mathilde Caron, Piotr Bojanowski, Julien Mairal, and Armand Joulin.
\newblock Unsupervised pre-training of image features on non-curated data.
\newblock In \emph{ICCV}, 2019.

\bibitem[Casanovas et~al.(2010)Casanovas, Monaci, Vandergheynst, and
  Gribonval]{casanovas2010blind}
Anna~Llagostera Casanovas, Gianluca Monaci, Pierre Vandergheynst, and R{\'e}mi
  Gribonval.
\newblock Blind audiovisual source separation based on sparse redundant
  representations.
\newblock \emph{IEEE Transactions on Multimedia}, 12\penalty0 (5):\penalty0
  358--371, 2010.

\bibitem[Chang et~al.(2017)Chang, Wang, Meng, Xiang, and Pan]{chang2017deep}
Jianlong Chang, Lingfeng Wang, Gaofeng Meng, Shiming Xiang, and Chunhong Pan.
\newblock Deep adaptive image clustering.
\newblock In \emph{ICCV}, pages 5879--5887, 2017.

\bibitem[Chen et~al.(2020{\natexlab{a}})Chen, Xie, Vedaldi, and
  Zisserman]{VGGSound}
Honglie Chen, Weidi Xie, Andrea Vedaldi, and Andrew Zisserman.
\newblock Vggsound: A large-scale audio-visual dataset.
\newblock \emph{ICASSP)}, May 2020{\natexlab{a}}.

\bibitem[Chen et~al.(2020{\natexlab{b}})Chen, Kornblith, Norouzi, and
  Hinton]{Chen2020ASF}
Ting Chen, Simon Kornblith, Mohammad Norouzi, and Geoffrey~E. Hinton.
\newblock A simple framework for contrastive learning of visual
  representations.
\newblock \emph{ArXiv}, abs/2002.05709, 2020{\natexlab{b}}.

\bibitem[Cho et~al.(2020)Cho, Kim, Chang, and Hwang]{cho2020selfsupervised}
Hyeon Cho, Taehoon Kim, Hyung~Jin Chang, and Wonjun Hwang.
\newblock Self-supervised spatio-temporal representation learning using
  variable playback speed prediction.
\newblock \emph{arXiv preprint arXiv:2003.02692}, 2020.

\bibitem[Christofides(1976)]{christofides1976worst}
Nicos Christofides.
\newblock Worst-case analysis of a new heuristic for the travelling salesman
  problem.
\newblock Technical report, Carnegie-Mellon Univ Pittsburgh Pa Management
  Sciences Research Group, 1976.

\bibitem[Cuturi(2013)]{sinkhornlightspeed}
Marco Cuturi.
\newblock Sinkhorn distances: Lightspeed computation of optimal transport.
\newblock In \emph{NeurIPS}, pages 2292--2300, 2013.

\bibitem[de~Sa(1994)]{NIPS1993_831}
Virginia~R. de~Sa.
\newblock Learning classification with unlabeled data.
\newblock In J.~D. Cowan, G.~Tesauro, and J.~Alspector, editors,
  \emph{NeurIPS}, pages 112--119. Morgan-Kaufmann, 1994.

\bibitem[Feo and Resende(1995)]{grasp3}
Thomas Feo and Mauricio Resende.
\newblock Greedy randomized adaptive search procedures.
\newblock \emph{Journal of Global Optimization}, 6:\penalty0 109--133, 03 1995.
\newblock \doi{10.1007/BF01096763}.

\bibitem[Festa and Resende(2004)]{festa2004annotated}
Paola Festa and Mauricio~GC Resende.
\newblock An annotated bibliography of grasp.
\newblock \emph{Operations Research Letters}, 8:\penalty0 67--71, 2004.

\bibitem[Gao et~al.(2018)Gao, Feris, and Grauman]{gao2018learning}
Ruohan Gao, Rogerio Feris, and Kristen Grauman.
\newblock Learning to separate object sounds by watching unlabeled video.
\newblock In \emph{ECCV}, pages 35--53, 2018.

\bibitem[Gidaris et~al.(2018)Gidaris, Singh, and
  Komodakis]{gidaris2018unsupervised}
Spyros Gidaris, Praveer Singh, and Nikos Komodakis.
\newblock Unsupervised representation learning by predicting image rotations.
\newblock \emph{ICLR}, 2018.

\bibitem[Gidaris et~al.(2020)Gidaris, Bursuc, Komodakis, Pérez, and
  Cord]{gidaris2020learning}
Spyros Gidaris, Andrei Bursuc, Nikos Komodakis, Patrick Pérez, and Matthieu
  Cord.
\newblock Learning representations by predicting bags of visual words, 2020.

\bibitem[Goyal et~al.(2017)Goyal, Doll{\'a}r, Girshick, Noordhuis, Wesolowski,
  Kyrola, Tulloch, Jia, and He]{goyal2017accurate}
Priya Goyal, Piotr Doll{\'a}r, Ross Girshick, Pieter Noordhuis, Lukasz
  Wesolowski, Aapo Kyrola, Andrew Tulloch, Yangqing Jia, and Kaiming He.
\newblock Accurate, large minibatch {SGD}: training imagenet in 1 hour.
\newblock \emph{arXiv preprint arXiv:1706.02677}, 2017.

\bibitem[Gu et~al.(2018)Gu, Sun, Ross, Vondrick, Pantofaru, Li,
  Vijayanarasimhan, Toderici, Ricco, Sukthankar, et~al.]{gu2018ava}
Chunhui Gu, Chen Sun, David~A Ross, Carl Vondrick, Caroline Pantofaru, Yeqing
  Li, Sudheendra Vijayanarasimhan, George Toderici, Susanna Ricco, Rahul
  Sukthankar, et~al.
\newblock Ava: A video dataset of spatio-temporally localized atomic visual
  actions.
\newblock In \emph{CVPR}, pages 6047--6056, 2018.

\bibitem[Han et~al.(2019)Han, Xie, and Zisserman]{han2019video}
Tengda Han, Weidi Xie, and Andrew Zisserman.
\newblock Video representation learning by dense predictive coding.
\newblock In \emph{ICCV}, 2019.

\bibitem[Hartigan(1972)]{hartigan1972direct}
John~A Hartigan.
\newblock Direct clustering of a data matrix.
\newblock \emph{Journal of the american statistical association}, 67\penalty0
  (337):\penalty0 123--129, 1972.

\bibitem[He et~al.(2016)He, Zhang, Ren, and Sun]{KaimingHe16}
Kaiming He, Xiangyu Zhang, Shaoqing Ren, and Jian Sun.
\newblock Deep residual learning for image recognition.
\newblock In \emph{CVPR}, 2016.

\bibitem[Hinton et~al.(2015)Hinton, Vinyals, and Dean]{hinton2015distilling}
Geoffrey Hinton, Oriol Vinyals, and Jeff Dean.
\newblock Distilling the knowledge in a neural network.
\newblock \emph{arXiv preprint arXiv:1503.02531}, 2015.

\bibitem[Hu et~al.(2019)Hu, Nie, and Li]{hu2019deep}
Di~Hu, Feiping Nie, and Xuelong Li.
\newblock Deep multimodal clustering for unsupervised audiovisual learning.
\newblock In \emph{CVPR}, pages 9248--9257, 2019.

\bibitem[Hu et~al.(2017)Hu, Miyato, Tokui, Matsumoto, and
  Sugiyama]{hu2017learning}
Weihua Hu, Takeru Miyato, Seiya Tokui, Eiichi Matsumoto, and Masashi Sugiyama.
\newblock Learning discrete representations via information maximizing
  self-augmented training.
\newblock In \emph{ICML}, pages 1558--1567, 2017.

\bibitem[Iakovidis et~al.(2018)Iakovidis, Georgakopoulos, Vasilakakis,
  Koulaouzidis, and Plagianakos]{iakovidis2018detecting}
Dimitris~K Iakovidis, Spiros~V Georgakopoulos, Michael Vasilakakis, Anastasios
  Koulaouzidis, and Vassilis~P Plagianakos.
\newblock Detecting and locating gastrointestinal anomalies using deep learning
  and iterative cluster unification.
\newblock \emph{IEEE transactions on medical imaging}, 37\penalty0
  (10):\penalty0 2196--2210, 2018.

\bibitem[Ji et~al.(2018)Ji, Henriques, and Vedaldi]{ji2018invariant}
Xu~Ji, João~F. Henriques, and Andrea Vedaldi.
\newblock Invariant information clustering for unsupervised image
  classification and segmentation, 2018.

\bibitem[Jiang et~al.(2019)Jiang, Zhao, Xia, Xue, Zhou, Ding, and
  Qian]{jiang2019novel}
Yizhang Jiang, Kaifa Zhao, Kaijian Xia, Jing Xue, Leyuan Zhou, Yang Ding, and
  Pengjiang Qian.
\newblock A novel distributed multitask fuzzy clustering algorithm for
  automatic mr brain image segmentation.
\newblock \emph{Journal of medical systems}, 43\penalty0 (5):\penalty0 118,
  2019.

\bibitem[Jing and Tian(2018)]{jing2018self}
Longlong Jing and Yingli Tian.
\newblock Self-supervised spatiotemporal feature learning by video geometric
  transformations.
\newblock \emph{arXiv preprint arXiv:1811.11387}, 2018.

\bibitem[Johnson et~al.(2017)Johnson, Douze, and J{\'e}gou]{faiss}
Jeff Johnson, Matthijs Douze, and Herv{\'e} J{\'e}gou.
\newblock Billion-scale similarity search with gpus.
\newblock \emph{arXiv preprint arXiv:1702.08734}, 2017.

\bibitem[Kay et~al.(2017)Kay, Carreira, Simonyan, Zhang, Hillier,
  Vijayanarasimhan, Viola, Green, Back, Natsev, Suleyman, and
  Zisserman]{kinetics}
Will Kay, Joao Carreira, Karen Simonyan, Brian Zhang, Chloe Hillier, Sudheendra
  Vijayanarasimhan, Fabio Viola, Tim Green, Trevor Back, Paul Natsev, Mustafa
  Suleyman, and Andrew Zisserman.
\newblock The kinetics human action video dataset.
\newblock \emph{CoRR}, abs/1705.06950, 2017.

\bibitem[Kim et~al.(2019)Kim, Cho, and Kweon]{kim2019self}
Dahun Kim, Donghyeon Cho, and In~So Kweon.
\newblock Self-supervised video representation learning with space-time cubic
  puzzles.
\newblock In \emph{AAAI}, 2019.

\bibitem[Korbar et~al.(2018)Korbar, Tran, and Torresani]{avts}
Bruno Korbar, Du~Tran, and Lorenzo Torresani.
\newblock Cooperative learning of audio and video models from self-supervised
  synchronization.
\newblock In \emph{NeurIPS}, 2018.

\bibitem[Kuehne et~al.(2011)Kuehne, Jhuang, Garrote, Poggio, and Serre]{HMDB51}
H.~Kuehne, H.~Jhuang, E.~Garrote, T.~Poggio, and T.~Serre.
\newblock {HMDB}: a large video database for human motion recognition.
\newblock In \emph{ICCV}, 2011.

\bibitem[Kuhn(1955)]{kuhn55hungarian}
Harold~W Kuhn.
\newblock The hungarian method for the assignment problem.
\newblock \emph{Naval research logistics quarterly}, 1955.

\bibitem[Lee et~al.(2017)Lee, Huang, Singh, and Yang]{lee2017unsupervised}
Hsin-Ying Lee, Jia-Bin Huang, Maneesh Singh, and Ming-Hsuan Yang.
\newblock Unsupervised representation learning by sorting sequences.
\newblock In \emph{ICCV}, 2017.

\bibitem[Lee et~al.(2019)Lee, Lee, and Teh]{lee2019deep}
Juho Lee, Yoonho Lee, and Yee~Whye Teh.
\newblock Deep amortized clustering.
\newblock \emph{arXiv preprint arXiv:1909.13433}, 2019.

\bibitem[Li et~al.(2020)Li, Zhou, Xiong, Socher, and Hoi]{li2020prototypical}
Junnan Li, Pan Zhou, Caiming Xiong, Richard Socher, and Steven~CH Hoi.
\newblock Prototypical contrastive learning of unsupervised representations.
\newblock \emph{arXiv preprint arXiv:2005.04966}, 2020.

\bibitem[Li and Wang(2020)]{li2020learning}
Tianhao Li and Limin Wang.
\newblock Learning spatiotemporal features via video and text pair
  discrimination, 2020.

\bibitem[Li et~al.(2019)Li, Dong, Wen, Hu, Zhou, and Zeng]{li2019clu}
Zhuoling Li, Minghui Dong, Shiping Wen, Xiang Hu, Pan Zhou, and Zhigang Zeng.
\newblock Clu-cnns: Object detection for medical images.
\newblock \emph{Neurocomputing}, 350:\penalty0 53--59, 2019.

\bibitem[Luo et~al.(2020)Luo, Liu, Zhou, Yang, Ma, Ye, and Wang]{luo2020video}
Dezhao Luo, Chang Liu, Yu~Zhou, Dongbao Yang, Can Ma, Qixiang Ye, and Weiping
  Wang.
\newblock Video cloze procedure for self-supervised spatio-temporal learning.
\newblock In \emph{AAAI}, 2020.

\bibitem[Miech et~al.(2019)Miech, Alayrac, Smaira, Laptev, Sivic, and
  Zisserman]{miech2019endtoend}
Antoine Miech, Jean-Baptiste Alayrac, Lucas Smaira, Ivan Laptev, Josef Sivic,
  and Andrew Zisserman.
\newblock End-to-end learning of visual representations from uncurated
  instructional videos, 2019.

\bibitem[Misra et~al.(2016)Misra, Zitnick, and Hebert]{misra2016shuffle}
Ishan Misra, C~Lawrence Zitnick, and Martial Hebert.
\newblock Shuffle and learn: unsupervised learning using temporal order
  verification.
\newblock In \emph{ECCV}, 2016.

\bibitem[Morgado et~al.(2020)Morgado, Vasconcelos, and Misra]{morgado2020avid}
Pedro Morgado, Nuno Vasconcelos, and Ishan Misra.
\newblock Audio-visual instance discrimination with cross-modal agreement,
  2020.

\bibitem[Nagrani et~al.(2020)Nagrani, Sun, Ross, Sukthankar, Schmid, and
  Zisserman]{nagrani2020}
Arsha Nagrani, Chen Sun, David Ross, Rahul Sukthankar, Cordelia Schmid, and
  Andrew Zisserman.
\newblock Speech2action: Cross-modal supervision for action recognition.
\newblock In \emph{CVPR}, 2020.

\bibitem[Noroozi and Favaro(2016)]{noroozi2016unsupervised}
Mehdi Noroozi and Paolo Favaro.
\newblock Unsupervised learning of visual representations by solving jigsaw
  puzzles.
\newblock In \emph{ECCV}, 2016.

\bibitem[Noroozi et~al.(2017)Noroozi, Pirsiavash, and
  Favaro]{noroozi2017representation}
Mehdi Noroozi, Hamed Pirsiavash, and Paolo Favaro.
\newblock Representation learning by learning to count.
\newblock In \emph{ICCV}, 2017.

\bibitem[Owens and Efros(2018)]{owens2018audio}
Andrew Owens and Alexei~A Efros.
\newblock Audio-visual scene analysis with self-supervised multisensory
  features.
\newblock In \emph{ECCV}, 2018.

\bibitem[Owens et~al.(2016)Owens, Wu, McDermott, Freeman, and
  Torralba]{owens2016ambient}
Andrew Owens, Jiajun Wu, Josh~H McDermott, William~T Freeman, and Antonio
  Torralba.
\newblock Ambient sound provides supervision for visual learning.
\newblock In \emph{ECCV}, 2016.

\bibitem[Patrick et~al.(2020)Patrick, Asano, Fong, Henriques, Zweig, and
  Vedaldi]{m2020multimodal}
Mandela Patrick, Yuki~M. Asano, Ruth Fong, João~F. Henriques, Geoffrey Zweig,
  and Andrea Vedaldi.
\newblock Multi-modal self-supervision from generalized data transformations,
  2020.

\bibitem[Piergiovanni et~al.(2020)Piergiovanni, Angelova, and
  Ryoo]{piergiovanni2020evolving}
AJ~Piergiovanni, Anelia Angelova, and Michael~S. Ryoo.
\newblock Evolving losses for unsupervised video representation learning, 2020.

\bibitem[Resende and Ribeiro(2019)]{resende2019greedy}
Mauricio~GC Resende and Celso~C Ribeiro.
\newblock Greedy randomized adaptive search procedures: Advances and
  extensions.
\newblock In \emph{Handbook of metaheuristics}, pages 169--220. Springer, 2019.

\bibitem[Rouditchenko et~al.(2019)Rouditchenko, Zhao, Gan, McDermott, and
  Torralba]{rouditchenko2019self}
Andrew Rouditchenko, Hang Zhao, Chuang Gan, Josh McDermott, and Antonio
  Torralba.
\newblock Self-supervised audio-visual co-segmentation.
\newblock In \emph{ICASSP}, 2019.

\bibitem[Senocak et~al.(2018)Senocak, Oh, Kim, Yang, and Kweon]{Senocak_2018}
Arda Senocak, Tae-Hyun Oh, Junsik Kim, Ming-Hsuan Yang, and In~So Kweon.
\newblock Learning to localize sound source in visual scenes.
\newblock \emph{CVPR}, Jun 2018.

\bibitem[Soomro et~al.(2012)Soomro, Zamir, and Shah]{UCF101}
Khurram Soomro, Amir~Roshan Zamir, and Mubarak Shah.
\newblock {UCF101}: A dataset of 101 human action classes from videos in the
  wild.
\newblock In \emph{CRCV-TR-12-01}, 2012.

\bibitem[Sun et~al.(2019{\natexlab{a}})Sun, Baradel, Murphy, and
  Schmid]{sun2019contrastive}
Chen Sun, Fabien Baradel, Kevin Murphy, and Cordelia Schmid.
\newblock Contrastive bidirectional transformer for temporal representation
  learning.
\newblock \emph{arXiv preprint arXiv:1906.05743}, 2019{\natexlab{a}}.

\bibitem[Sun et~al.(2019{\natexlab{b}})Sun, Myers, Vondrick, Murphy, and
  Schmid]{sun2019videobert}
Chen Sun, Austin Myers, Carl Vondrick, Kevin Murphy, and Cordelia Schmid.
\newblock Videobert: A joint model for video and language representation
  learning.
\newblock In \emph{ICCV}, pages 7464--7473, 2019{\natexlab{b}}.

\bibitem[Tian et~al.(2018)Tian, Shi, Li, Duan, and Xu]{tian2018ave}
Yapeng Tian, Jing Shi, Bochen Li, Zhiyao Duan, and Chenliang Xu.
\newblock Audio-visual event localization in unconstrained videos.
\newblock In \emph{ECCV}, 2018.

\bibitem[Tian et~al.(2019)Tian, Krishnan, and Isola]{tian2019contrastive}
Yonglong Tian, Dilip Krishnan, and Phillip Isola.
\newblock Contrastive multiview coding.
\newblock \emph{arXiv preprint arXiv:1906.05849}, 2019.

\bibitem[Tran et~al.(2018)Tran, Wang, Torresani, Ray, LeCun, and
  Paluri]{Tran18}
Du~Tran, Heng Wang, Lorenzo Torresani, Jamie Ray, Yann LeCun, and Manohar
  Paluri.
\newblock A closer look at spatiotemporal convolutions for action recognition.
\newblock In \emph{CVPR}, 2018.

\bibitem[Ulyanov et~al.(2017)Ulyanov, Vedaldi, and
  Lempitsky]{ulyanov17deepimageprior}
Dmitry Ulyanov, Andrea Vedaldi, and Victor Lempitsky.
\newblock Deep image prior.
\newblock \emph{arXiv preprint arXiv:1711.10925}, 2017.

\bibitem[Van~Gansbeke et~al.(2020)Van~Gansbeke, Vandenhende, Georgoulis,
  Proesmans, and Van~Gool]{gansbeke2020learning}
Wouter Van~Gansbeke, Simon Vandenhende, Stamatios Georgoulis, Marc Proesmans,
  and Luc Van~Gool.
\newblock Scan: Learning to classify images without labels.
\newblock In \emph{European Conference on Computer Vision (ECCV)}, 2020.

\bibitem[Wang et~al.(2019)Wang, Jiao, Bao, He, Liu, and Liu]{motion_statistics}
Jiangliu Wang, Jianbo Jiao, Linchao Bao, Shengfeng He, Yunhui Liu, and Wei Liu.
\newblock Self-supervised spatio-temporal representation learning for videos by
  predicting motion and appearance statistics.
\newblock In \emph{CVPR}, 2019.

\bibitem[Wei et~al.(2018)Wei, Lim, Zisserman, and Freeman]{wei2018learning}
Donglai Wei, Joseph~J Lim, Andrew Zisserman, and William~T Freeman.
\newblock Learning and using the arrow of time.
\newblock In \emph{CVPR}, 2018.

\bibitem[Wu et~al.(2018)Wu, Xiong, Yu, and Lin]{Wu_2018_CVPR}
Zhirong Wu, Yuanjun Xiong, Stella~X. Yu, and Dahua Lin.
\newblock Unsupervised feature learning via non-parametric instance
  discrimination.
\newblock In \emph{CVPR}, June 2018.

\bibitem[Xiao et~al.(2020)Xiao, Lee, Grauman, Malik, and
  Feichtenhofer]{xiao2020audiovisual}
Fanyi Xiao, Yong~Jae Lee, Kristen Grauman, Jitendra Malik, and Christoph
  Feichtenhofer.
\newblock Audiovisual slowfast networks for video recognition.
\newblock \emph{arXiv preprint arXiv:2001.08740}, 2020.

\bibitem[Xie et~al.(2016)Xie, Girshick, and Farhadi]{xie2016unsupervised}
Junyuan Xie, Ross Girshick, and Ali Farhadi.
\newblock Unsupervised deep embedding for clustering analysis.
\newblock In \emph{International conference on machine learning}, pages
  478--487, 2016.

\bibitem[Xu et~al.(2019)Xu, Xiao, Zhao, Shao, Xie, and Zhuang]{clip_order}
Dejing Xu, Jun Xiao, Zhou Zhao, Jian Shao, Di~Xie, and Yueting Zhuang.
\newblock Self-supervised spatiotemporal learning via video clip order
  prediction.
\newblock In \emph{CVPR}, 2019.

\bibitem[Yan et~al.(2020)Yan, Misra, Gupta, Ghadiyaram, and
  Mahajan]{yan2020cluster}
Xueting Yan, Ishan Misra, Abhinav Gupta, Deepti Ghadiyaram, and Dhruv Mahajan.
\newblock {ClusterFit: Improving Generalization of Visual Representations}.
\newblock In \emph{CVPR}, 2020.

\bibitem[Yang et~al.(2016)Yang, Parikh, and Batra]{yang2016joint}
Jianwei Yang, Devi Parikh, and Dhruv Batra.
\newblock Joint unsupervised learning of deep representations and image
  clusters.
\newblock In \emph{CVPR}, pages 5147--5156, 2016.

\bibitem[Zhang et~al.(2016)Zhang, Isola, and Efros]{zhang2016colorful}
Richard Zhang, Phillip Isola, and Alexei~A Efros.
\newblock Colorful image colorization.
\newblock In \emph{ECCV}, 2016.

\bibitem[Zhao et~al.(2019)Zhao, Gan, Ma, and Torralba]{zhao2019sound}
Hang Zhao, Chuang Gan, Wei-Chiu Ma, and Antonio Torralba.
\newblock The sound of motions.
\newblock In \emph{ICCV}, pages 1735--1744, 2019.

\end{thebibliography}

\def\optrow#1\\{#1\\} %
\def\labelswitch#1{\label{supp:#1}}
\def\dontshowthisinappendix#1{}
\def\showthisinappendix#1{#1}

\newpage
\appendix
\counterwithin{figure}{section}
\counterwithin{table}{section}
\section{Appendix}
\subsection{Pretrained model details \label{appx:others-input-details}}

Here we provide additional information about the pretrained models we have used in this work.
\begin{table}[!htb]
    \centering
    \caption{\textbf{Details for audio encoder.} Architectural and pretraining details for XDC's audio encoder used for benchmarking. \label{tab:details-audio}}
    \vspace{1em}
    \begin{tabular}{l l l l}
    \toprule
        Method & Input shape          & Architecture & Pretrain dataset \\
         \midrule
        XDC    & $40\times1\times100$ & Resnet-18       & Kinetics-400\\
    \bottomrule
    \end{tabular}
\end{table}

\subsection{Implementation details \label{appx:training-details}}
We train our method using the Sinkhorn-Knopp parameter $\lambda=20$, an inverse quadratic clustering schedule with $100$ clustering operations and 10 heads which we adopt from \citep{asano2020self}.
For evaluation, we report results for head 0 to compare against the ground-truth, as we found no significant difference in performance between heads.
For the Gaussian distribution, we take the marginals to be from $\mathcal{N}(1, 0.1)*N/K$.
For the clustering-heads, we use two-layer MLP-heads as in \citep{Chen2020ASF,bachman2019learning}.
The video inputs are 30 frame long clips sampled consecutively from 30fps videos and are resized such that the shorter side is 128 and during training a random crop of size 112 is extracted, no color-jittering is applied. Random horizontal flipping is applied to the video frames with probability $0.5$, and then the channels of the video frames are Z-normalized using mean and standard deviation statistics computed across the dataset. 
The audio is processed as a $1\times257\times199$ image, by taking the log-mel bank features with 257 filters and 199 time-frames and for training, random volume jittering between 90\% and 110\% is applied to raw waveform, similar to~\citep{morgado2020avid}.
For evaluation, a center-crop is taken instead for the video inputs and audio volume is not jittered.
We use a mini-batch size of $16$ on each of our $64$ GPUs giving an effective batch size of $1024$ for distributed training for 200 epochs.
The initial learning rate is set to $0.01$ which we linearly scale with the number of GPUs, after following a gradual warm-up schedule for the first $10$ epochs~\citep{goyal2017accurate}.
For training on Kinetics-Sound and AVE, we initialize our model with a VGG-Sound pretrained backbone due to the small training set sizes ($N=22$k and $N=3328$).
The clustering heads are re-initialized randomly.
This ensures a more fair comparison as XDC, DPC and the supervised model are pretrained on Kinetics-400 with $N=230$k and MIL-NCE on HowTo100M with $N=100$M videos.
We train on VGG-Sound for 200 epochs, which takes around 2 days.

\subsection{Pair-based optimization for AV-Alignment\label{appx:pairwise-details}}
For aligning the visual and audio encoder, we use a greedy switching algorithm that starts from a feasible initial solution~\citep{resende2019greedy,festa2004annotated,grasp3}.
In particular, we consider 50000 potential pair switches with 5 randomized restarts and take the final permutation that yields the lowest cost.

\subsection{Evaluation metrics details \label{appx:entropy-purity}}
The \textbf{normalized mutual information} (NMI) is calculated by the formula
\begin{equation}
    \text{NMI} = \frac{\text{MI}(U,V)}{0.5H(U) +0.5H(V)},
\end{equation}
where the Mutual information $\mathrm{MI}$ is given by 
$\text{MI}(U, V) = \sum_{i=1}^{|U|}\sum_{j=1}^{|V|}P(i, j)\log\left(\frac{P(i,j)}{P(i)P'(j)}\right)$, and $H$ is the standard entropy, with $H(U) = - \sum_{i=1}^{|U|}P(i)\log(P(i))$.
The NMI ranges from 0 (no mutual information) to 100\%, which implies perfect correlation.

The rand index (RI) is given by $\text{RI} = \frac{a + b}{C}$, where $a,b$ are the number of pairs of elements that are in the same/different set in the ground truth labelling and in the same/different set in the predicted clustering and $C$ is the total number of such pairs.
The a\textbf{djusted Rand index} (ARI) corrects for random assignments and is given by 
\begin{equation}
    \text{ARI} = \frac{\text{RI} - E[\text{RI}]}{\max(\text{RI}) - E[\text{RI}]},
\end{equation}
where the expected RI of a random label assignment is subtracted in both nominator and denominator.
Due to the subtraction, the ARI varies from -1 to 1 with a value close to 0 implying random correlation and a value of 1 implying identical agreement.

The \textbf{mean entropy per cluster} is given by
\begin{equation}
    \langle H \rangle = \frac{1}{K}\sum_{k \in K} H(p(y \vert \hat{y}_k = k)), \label{eq:entropy}
\end{equation}
where $\hat{y}$ are unsupervisedly obtained clusters and $p(y \vert \hat{y}_k = k)$ is the distribution of ground-truth clusters for cluster $k$. 
Hence, the optimal number of this metric is $0$ and a chance assignment yields $\langle H \rangle = -\log{1/K}$. %

Further, as we wish to understand the semantic purity compared to the ground truth labels of each cluster, so we additionally report the the \textbf{mean maximal purity per cluster},
\begin{equation}
    \langle p_\mathrm{max} \rangle = \frac{1}{K}\sum_{k \in K} \max(p(y \vert \hat{y}_k = k)), \label{eq:purity}
\end{equation}
which ranges from $\langle p_\mathrm{max} \rangle =1/K$ (chance level) to perfect matching at $ \langle p_\mathrm{max} \rangle =100\%$.

\subsection{Single modality degradation experiment details\label{degraded-detail}}
We use the default input-sizes for each model, i.e. 112 for ours and the supervised model, 224 for MIL-NCE.
Compression is implemented by nearest-neighbor downsampling and subsequently nearest-neighbor upsamling for speed. 
For this experiment only, we evaluate the performance on the smaller validation sets.

\subsection{Further ablations\label{further-ablation}}

In \cref{tab:appx:numcluster}, we provide the results for varying the number of clusters $K$ in our algorithm.
We find that even when moving from the ground-truth number of classes ($K=309$), to lower numbers ($K=256$) or higher estimates ($K=619,1024$) our results remain stable with the NMI staying almost constant.
While the ARI does drop for larger $K$, we also observe an increase in the purity of the clusters for a larger number of clusters from $\langle \mathbf{p_\mathrm{max}} \rangle=38.0$ for $K=309$ to $\langle \mathbf{p_\mathrm{max}} \rangle=42.7$ for $K=619$, which can be particularly useful when dividing the dataset into clusters and subsequently only obtaining human annotations for few examples per cluster.
\begin{table}[htb]
\setlength{\tabcolsep}{3pt}
   \centering
   \footnotesize
   \caption{\textbf{Varying $\mathbf{K}$ } in our method degrades performances only slightly, showing that our method is robust to various estimations of the ground-truth number of classes. Results on VGG-Sound. \label{tab:appx:numcluster}}
   \vspace{1em}
  \begin{tabular}{l l @{\hskip 5pt} c c c c c}
  \toprule
     Method                        & $K$ &  \textbf{NMI} & \textbf{ARI} & \textbf{Acc.} & $\langle \mathbf{H} \rangle$ & $\langle \mathbf{p_\mathrm{max}} \rangle$ \\
    \midrule 
    \textbf{SeLaVi} & 309 &  56.7 & 22.5 & 32.3 & 2.4 & 38.0 \\
    \midrule
    \textbf{SeLaVi} & 256 &  56.8 & 24.3 & 34.2 & 2.4 & 36.9 \\
    \textbf{SeLaVi} & 619 &  56.9 & 16.8 & 23.0 & 2.2 & 42.7 \\
    \textbf{SeLaVi} & 1024 & 55.1 & 16.3 & 9.6  & 2.1  & 42.2  \\
    \bottomrule
  \end{tabular}

\end{table}

\subsection{Retrieval downstream task implementation details}
We follow~\citep{clip_order} in our evaluation protocol and use split 1 of UCF101 and HMDB-51.  
We uniformly sample 10 clips per video, and average the max-pooled features after the last residual block for each clip per video.
We then utilize the averaged features from the validation set to query the videos in the training set.
The cosine distance of representations between the query clip and all clips in the training set are computed and when the class of a test clip appears in the classes of k nearest training clips, it is considered to be correctly retrieved.
R@$k$ refers to the retrieval performance using $k$ nearest neighbors.

\subsection{Visual classification downstream task}
\begin{table}[!ht]
	\centering
	\tabcolsep=0.09cm
	\caption{\textbf{Representation learning downstream evaluation.}
	Self-supervised and fully-supervised trained methods on UCF101 and HMDB51 benchmarks.
	We follow the standard protocol and report the average top-1 accuracy over the official splits and show results for finetuning the whole network.
	Methods with $^\dagger$ indicate the additional use of video titles and ASR generated text as supervision.
	Methods with $^*$ use ASR generated text.
	\label{tab:visual-downstream} }
	\vspace{1em}
	\begin{tabular}{l l l @{\hskip 0.2in} c c }
		\toprule
		\textbf{Method}                              & \textbf{Architecture} & \bf{Pretrain Dataset}  & %
		\multicolumn{2}{c}{\underline{\textbf{Top-1 Acc\%}}} \\
		                                             &                       &                        & UCF\,            & HMDB \\
		\midrule
		Full supervision~\citep{alwassel2019self}                             & R(2+1)D-18            & ImageNet               & 82.8             & 46.7\\
		Full supervision~\citep{alwassel2019self}                             & R(2+1)D-18            & Kinetics-400           & \bf{93.1} & \bf{63.6} \\
		Weak supervision,
		CPD~\citep{li2020learning}$^\dagger$         & 3D-Resnet50           & Kinetics-400& 88.7             & 57.7 \\
		\midrule
        MotionPred~\citep{motion_statistics}         & C3D                   & Kinetics-400           & 61.2             & 33.4 \\
		RotNet3D~\citep{jing2018self}                & 3D-ResNet18           & Kinetics-600           & 62.9             & 33.7 \\
		ST-Puzzle~\citep{kim2019self}                & 3D-ResNet18           & Kinetics-400           & 65.8             & 33.7 \\
		ClipOrder~\citep{clip_order}                 & R(2+1)D-18            & Kinetics-400                    & 72.4                & 30.9 \\
		DPC~\citep{han2019video}                     & 3D-ResNet34           & Kinetics-400           & 75.7             & 35.7 \\
		CBT~\citep{sun2019contrastive}               & S3D                   & Kinetics-600           & 79.5             & 44.6   \\
		Multisensory~\citep{owens2018audio}          & 3D-ResNet18           & Kinetics-400           & 82.1             & - \\
		XDC~\citep{alwassel2019self}                 & R(2+1)D-18            & Kinetics-400           & 84.2             & 47.1 \\
		AVTS~\citep{avts}                            & MC3-18                & Kinetics-400           & 85.8             & 56.9  \\
		AV Sync+RotNet~\citep{xiao2020audiovisual}   & AVSlowFast            & Kinetics-400           & 87.0             & 54.6 \\
		GDT  ~\citep{m2020multimodal}                & R(2+1)D-18            & Kinetics-400           & \ul{88.7}    & \ul{57.8} \\
		\midrule
		\bf{\methodname}                      & R(2+1)D-18            & Kinetics-400           & 83.1   & 47.1 \\
		\bf{\methodname}                      & R(2+1)D-18            & VGG-Sound              & 87.7    & 53.1 \\

		\bottomrule
	\end{tabular}
	\vspace{2mm}

\end{table}

In \Cref{tab:visual-downstream} we show the performance of our method on two common visual-only video feature representation benchmarks, UCF-101~\citep{UCF101} and HMDB-51~\cite{HMDB51}.
Note that, as is the standard in this evaluation, we use our visual encoder as initialization and fine-tune the whole network on the target down-stream task.
In particular, we follow the finetuning schedule of the one of the current state-of-the-art methods~\citep{m2020multimodal}.
We find that we achieve competitive performance when trained on VGG-Sound, even surpassing XDC, despite our method using only a spatial resolution of $112\times112$ and not $224\times224$.

\end{document}